\documentclass[lettersize,journal]{IEEEtran}
\usepackage{amsmath,amsfonts}
\usepackage{algorithmic}
\usepackage{algorithm}
\usepackage{array}
\usepackage[caption=false,font=normalsize,labelfont=sf,textfont=sf]{subfig}
\usepackage{textcomp}
\usepackage{stfloats}
\usepackage{url}
\usepackage{verbatim}
\usepackage{graphicx}
\usepackage{soul}
\usepackage{graphicx}

\usepackage{tikz}
\usepackage{comment}
\usepackage{amsmath,amssymb} 
\usepackage{color}
\soulregister\cite7
\soulregister\ref7
\usepackage[accsupp]{axessibility}  

\usepackage[misc]{ifsym}
\usepackage{epsfig}
\usepackage{type1cm}
\usepackage{stfloats}
\usepackage{booktabs}
\usepackage{multirow}
\usepackage{multicol}
\usepackage{lineno}
\usepackage{bm}
\usepackage{threeparttable}
\usepackage{cite}
\makeatletter
\let\NAT@parse\undefined \makeatother
\usepackage[pagebackref=true,breaklinks=true,colorlinks,bookmarks=false]{hyperref}
\usepackage{subfig}
\usepackage{makecell}
\newcommand{\ie}{\textit{i}.\textit{e}., }

\begin{document}

\title{Uncertainty-Driven Action Quality Assessment}

\author{Caixia~Zhou and Yaping~Huang\IEEEauthorrefmark{1}
	\thanks{ Caixia~Zhou and Yaping Huang are with the Beijing Key Laboratory of Traffic Data Analysis and Mining, Beijing Jiaotong University, Beijing 100044, China (e-mail: cxzhou@bjtu.edu.cn; yphuang@bjtu.edu.cn).}
\thanks{\textit{Corresponding authors: Yaping Huang.}}
}

\markboth{Journal of \LaTeX\ Class Files,~Vol.~14, No.~8, August~2021}%
{Shell \MakeLowercase{\textit{et al.}}: A Sample Article Using IEEEtran.cls for IEEE Journals}
\maketitle

\begin{abstract}
Automatic action quality assessment (AQA) has attracted increasing attention due to its wide applications. However, most existing AQA methods employ deterministic models to predict the final score for each action, while overlooking the subjectivity and diversity among expert judges during the scoring process. 
In this paper, we propose a novel probabilistic model, named Uncertainty-Driven AQA (UD-AQA), to utilize and capture the diversity among multiple judge scores. 
Specifically, we design a Conditional Variational Auto-Encoder (CVAE)-based module to encode the uncertainty in expert assessment, where multiple judge scores can be produced by sampling latent features from the learned latent space multiple times. To further utilize the uncertainty, we generate the estimation of uncertainty for each prediction, which is employed to re-weight AQA regression loss, effectively reducing the influence of uncertain samples during training. Moreover, we further design an uncertainty-guided training strategy to dynamically adjust the learning order of the samples from low uncertainty to high uncertainty. The experiments show that our proposed method achieves competitive results on three benchmarks including the Olympic events MTL-AQA and FineDiving, and the surgical skill JIGSAWS datasets. 
\end{abstract}


\section{Introduction}
\IEEEPARstart{T}{he} task of action quality assessment (AQA) aims to grade the execution of an action, such as in sport scoring~\cite{Tang2020uncertainty5, Nekoui2021EAGLE6, Parmar2019action7}, surgical skill assessment~\cite{gao2014jhu40,funke2019video,lavanchy2021automation,wang2020towards}, piano skills assessment~\cite{Parmar2021piano33}, and skill determination in daily life~\cite{Doughty2019the34,doughty2021skill35,li2019manipulation}.  It is common to invite multiple judges\footnote{In this paper, the term ``judge" refers to a human expert who evaluates an action video and provides an assessment score.} to score an action for fairness and comprehensiveness. Study of automatic AQA scoring system~\cite{Pirsiavash2014assessing1, parmar2017learning2, Wang2020assessing3, Parmar2019what4, Tang2020uncertainty5, Nekoui2021EAGLE6, Parmar2019action7, Pan2019Action8} has attracted increasing interests due to its objectivity and labor-saving. In contrast to action recognition tasks that involve significantly different action categories~\cite{wang2016temporal36,wang2021action,yang2020temporal,li2020tea,mou2023compressed,shen2021fexnet}, the videos used in AQA analysis typically consist of hundreds of frames where actions and environments exhibit subtle variations.  This inherent similarity between actions and environments poses a greater challenge in distinguishing various scores, emphasizing the need to capture the subtle variations that contribute to the distinctions between videos.
\begin{figure}[ht]
 \centering
 \includegraphics[scale=0.42]{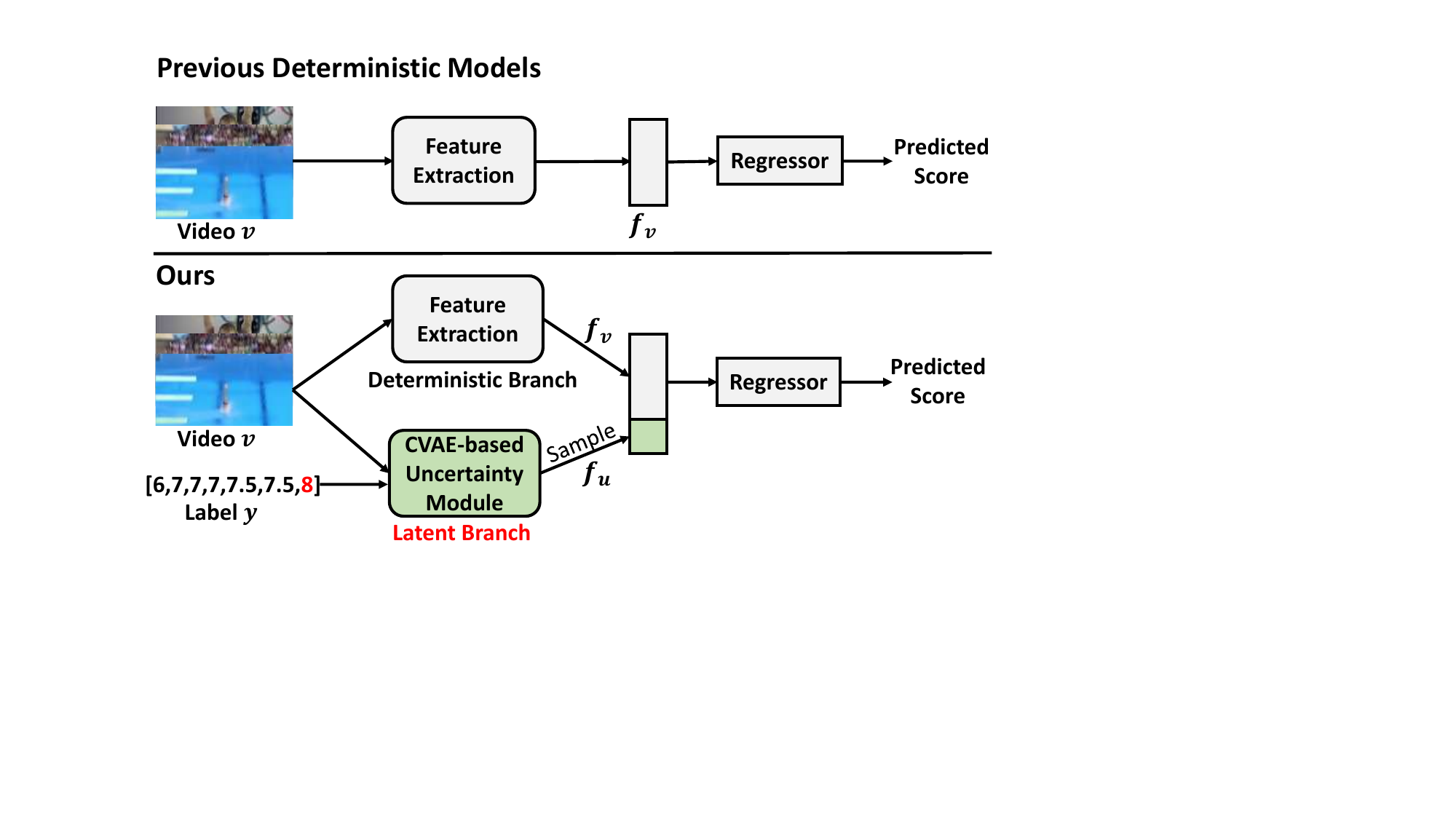}
 \caption{{Previous deterministic models and our proposed probabilistic UD-AQA model. The top figure illustrates the limitations of previous deterministic models, which can only predict one deterministic result. In contrast, the bottom figure showcases our proposed UD-AQA model with a novel latent branch encoding uncertainty among judge scores. UD-AQA combines latent features and deterministic features to generate prediction. By sampling multiple times, UD-AQA can produce diverse results.}}
 \label{Fig1}
\end{figure}

To extract distinguishable features from action videos, some methods directly learn video-level representations through 3D convolution~\cite{Tang2020uncertainty5,zhang2022semi,zhou2023hierarchical}, while others obtain frame-level representations through 2D convolution first and subsequently incorporate temporal information using 1D convolution~\cite{li2022pairwise,lian2023improving}. Some studies introduce pose-based representations~\cite{pan2021adaptive,dong2021learning} which first recognize the sub-actions in the videos and then extract related features. After generating action representations, the majority of existing methods evaluate action quality using three learning strategies: classification-based~\cite{Tang2020uncertainty5}, regression-based~\cite{Parmar2019what4, Pan2019Action8}, and contrastive learning-based~\cite{yu2021group42, bai2022action, li2022pairwise}.

All methods mentioned above are deterministic models, predicting only a single final score for an action. However, such deterministic models overlook the inherent ambiguity of the AQA task. To give a fair result, multiple judges are invited to score the actions, which are available in the AQA dataset, making it important to enable the learned model to emulate these multiple judges. However, most current methods ignore these multiple scores or simply construct multi-head models for prediction~\cite{Tang2020uncertainty5}. 
Although these methods have achieved promising performance, they are expensive and lack flexibility in handling scenarios with a variable number of scores. If more scores are needed, the multi-head model must be rebuilt and re-trained.  

To address the mentioned issues, in this paper, we propose an Uncertainty-Driven Action Quality Assessment (UD-AQA) model, which is designed to encode the uncertainty arising from ambiguity among judges. Generative models can provide multiple potential solutions and learn various patterns in the presence of ambiguous or fuzzy inputs, thus motivated,
we introduce a Conditional Variational Auto-Encoder (CVAE)-based~\cite{sohn2015learning39} uncertainty estimation module which can effectively capture the inherent ambiguity in the AQA task. The proposed UD-AQA can sample multiple times from the latent space, generating diverse predictions that emulate the scoring process from different judges. As a result, it can provide a variable number of scores. Moreover, the uncertainty estimation module enhances the utilization of judge scores, contributing to a more accurate prediction of the final score.
Figure~\ref{Fig1} illustrates the distinction between previous deterministic models and the proposed probabilistic UD-AQA.

Specifically, the proposed UD-AQA consists of two branches: the deterministic branch and the latent branch. The deterministic branch is employed to obtain deterministic features of the action videos, while the latent branch is used to model the inherent ambiguity of the action videos. In the latent branch, UD-AQA constructs a latent space learned from the given video and its corresponding scores through CVAE. Then we combine the video-level features extracted from the Inflated 3D ConvNets (I3D)~\cite{Carreira2017Quo23} with latent features sampled from the CVAE module to generate fused representations. 
During the training process, we randomly sample one score from the judge score sets as the score label each time. During the test process, we perform sampling multiple times from the CVAE-based module to generate diverse results. The number of sampling is determined by the required number of judges. The random sampling from the CVAE-based module introduces rich diversity, allowing for the generation of multiple scores.

Moreover, we further map the multi-dimensional latent space into a one-dimensional scalar to quantitatively estimate uncertainty.
The estimated uncertainty is then used to re-weight the AQA regression loss. The intuition is that certain samples should carry higher weights, whereas uncertain ones should have lower importance during training.
We further propose a novel training scheme, which decides the training order of samples dynamically based on the predicted uncertainty. This concept draws inspiration from curriculum learning~\cite{bengio2009curriculum} and self-paced learning~\cite{kumar2010self}, which suggest that a learning model can perform better when examples are organized in an easy-to-hard order. Motivated by this, we sort the samples from low to high uncertainty, allowing the model to prioritize learning from higher-confidence information first. The experiments on three AQA datasets show that UD-AQA achieves competitive performance, especially achieving a new state-of-the-art on the recent large-scale FineDiving dataset. 

In summary, our contributions can be summarized as follows:

\begin{itemize}
    \item We propose a novel probabilistic model, named Uncertainty-Driven AQA (UD-AQA), to capture and exploit the ambiguity among multiple judge scores via a CVAE-based uncertainty estimation module, which is capable of generating multiple predictions through sampling from the constructed latent space. 
    \item We utilize the estimated uncertainty to re-weight the AQA regression loss. The weight is higher when uncertainty is low, explicitly reducing the contributions of uncertain samples during training.
    \item We design a novel uncertainty-guided training strategy that organizes the training samples based on the predicted uncertainty. This ensures that the model learns high-confidence information first.
    \item We conduct comprehensive experiments on three AQA benchmarks. Our proposed UD-AQA achieves competitive performance, especially on the recently proposed large-scale FineDiving dataset.
\end{itemize}

\section{Related Work}
\subsection{Action Quality Assessment}
Assessing the quality of actions automatically has been explored for a long time in the computer vision community. The first learning-based attempt~\cite{Pirsiavash2014assessing1} uses hand-crafted features to train a Support Vector Regression (SVR) model, which regresses the scores of input actions. Recently, deep learning-based methods~\cite{parmar2017learning2} have demonstrated promising performance~\cite{Wang2020assessing3, Parmar2019what4, Tang2020uncertainty5, Nekoui2021EAGLE6, yu2021group42}. These methods typically employ deep neural networks to generate distinctive feature representations. They segment action videos into several parts to extract clip-level features, which are then aggregated into video-level features.

\begin{figure*}[ht]
 \centering
 \includegraphics[scale=0.55]{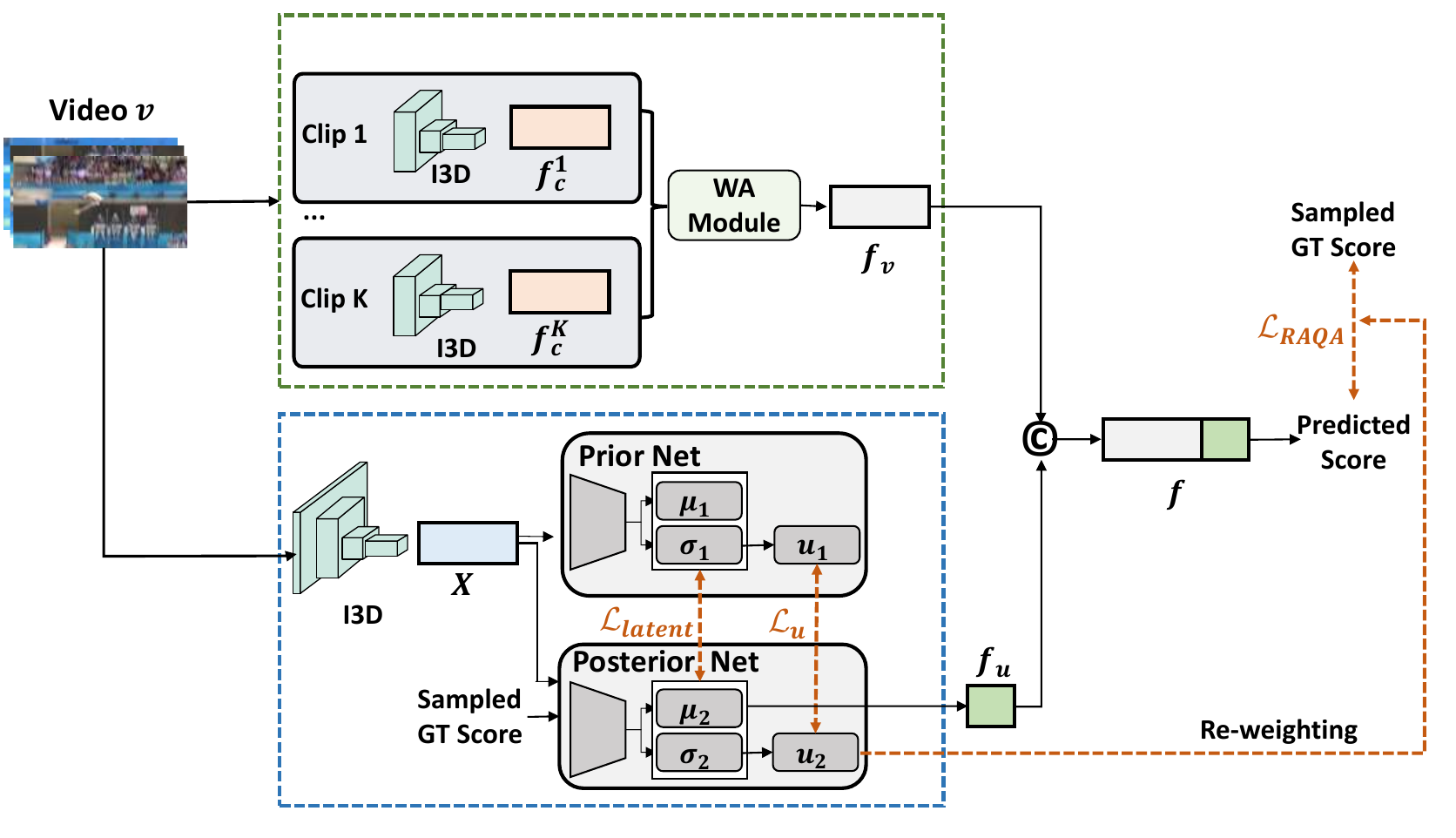}
 \caption{{The overall framework of our proposed UD-AQA. The input videos are segmented into $K$ overlapping clips and fed into the I3D backbone to extract the clip-level features. Then we design a weight attention (WA) module to aggregate the clip-level features into video-level features. To model ambiguity among judges, we propose a CVAE-based module that projects each video and its corresponding judge scores into a low-dimensional latent space, allowing us to sample latent variables and produce multiple outputs. Dashed lines and boxes indicate components used only in training.}}
 \label{Fig2}
\end{figure*}

Parmar and Morris~\cite{parmar2017learning2} employ C3D~\cite{tran2015learning43} to extract clip-level features, and the average pooling or Long Short-Term Memory (LSTM)~\cite{hochreiter1997long} are used to generate the video-level features. Fully Connected (FC) layers or SVR are utilized to predict the final scores.~\cite{Parmar2019action7} and~\cite{Parmar2019what4} also use C3D to generate features, and ~\cite{Parmar2019what4} trains the model across actions. Differently,~\cite{Parmar2019action7} uses multi-task learning to predict action, score, and caption simultaneously.  MUSDL~\cite{Tang2020uncertainty5} chooses Inflated 3D ConvNets (I3D)~\cite{Carreira2017Quo23} to extract the features, and adopts several FC layers to produce the scores. \cite{Wang2020assessing3} implements a spatial and temporal convolutional network to extract features at a full-frame rate and utilizes an attention mechanism to fuse the features in the temporal dimension. 

Some works introduce human pose information to assess the quality of actions.~\cite{Pirsiavash2014assessing1} uses spatial-temporal Gabor-like filters~\cite{le2011learning} to capture low-level features and applies Discrete Fourier Transform on the pose sequence. ~\cite{Pan2019Action8} explores temporal and spatial relation graphs on extracted joint motion features which are obtained from local patches around joints' I3D features. Then, temporal difference, spatial difference, motion, and whole-scene features are used together to get scores. EAGLE-Eye~\cite{Nekoui2021EAGLE6} trains an extreme pose estimator using the HRNet~\cite{Sun2019deep32} on the G-ExPose dataset~\cite{Nekoui2021EAGLE6}, which can capture pose and appearance features to reason the coordination among the joints and appearance dynamics.

The existing state-of-the-art methods are built upon the contrastive learning~\cite{chen2020simple,he2020momentum} strategy. PLCN~\cite{li2022pairwise} designs consistency constraints to learn relative scores between two videos. CoRe~\cite{yu2021group42} and TQT~\cite{bai2022action} use exemplar videos and predict relative scores by constructing a group-aware regression tree (GART). TQT also utilizes a transformer-based structure to parse the extracted video-level features into multiple temporally ordered key phases. Both CoRe and TQT apply 10 exemplar videos for testing to ensure the superiority of performance.

Unlike the aforementioned methods, we believe that judge scores are \textcolor{red}{objective} yet contain uncertain factors. Therefore, we incorporate uncertainty into the network to predict multiple plausible scores. While MUSDL~\cite{Tang2020uncertainty5} also defines an uncertainty-aware regression model, it primarily focuses on softening single labels into Gaussian distributions to obtain the finer-level score annotations. In contrast, the uncertainty in our work is primarily aimed at modeling the ambiguity inherent in expert assessment of the AQA task.
This type of uncertainty is extensively studied in the field of uncertainty estimation but has not been explored in the context of the AQA task.
Specifically, we encode the uncertainty via CVAE to generate multiple scores for each action. Moreover, the predicted uncertainty is also useful for re-weighting the training loss and organizing the training order of the samples.

\subsection{Conditional Variational Auto-Encoder} 
CVAE~\cite{sohn2015learning39} is a popular deep conditional generative model~\cite{kingma2013auto} designed for structured output prediction using Gaussian latent spaces. Generally, a CVAE consists of a prior network, a posterior network, and a generation network. The variables in the CVAE model include input variables, output variables, and latent variables. Unlike the VAE model whose prior is a standard Gaussian distribution, the prior of the CVAE model is a learnable Gaussian distribution related to the input variables. The posterior of the CVAE model is a Gaussian distribution related to input variables and conditions. The training goal is to narrow the distributions of the prior network and posterior network, pushing the prediction of the generation network close to the ground truth.

With latent variables, CVAE possesses exclusive generative ability, which can be applied to generate diverse outputs~\cite{sohn2015learning39}. Probabilistic UNet~\cite{Kohl2018probabilistic20}, PhiSeg~\cite{Baumgartner2019PhiSeg21}, and RevPhiseg~\cite{gantenbein2020revphiseg45} incorporate CVAE in the medical image segmentation task, where erroneous predictions pose significant risks. Thus, having multiple results for reference becomes necessary. CVID~\cite{du2020conditional46} harnesses CVAE to provide diverse predictions for rainy images and then averages these predictions to enhance model performance. UCNet~\cite{Zhang2020UCnet13, zhang2021uncertainty44} employs an adaptive threshold and majority voting strategy to select the final predictions for the saliency detection task.
 
To the best of our knowledge, CVAE has not been exploited in the AQA task. We are the first to employ CVAE in the AQA task, taking into account expert assessment uncertainty. By combining features from visual backbones and features sampled from the latent space, the model can produce different and diverse scores for AQA tasks, thereby further capturing the ambiguity and improving the performance.

\section{Proposed Approach}
In this section, we introduce the proposed UD-AQA, which builds on the intuition that multiple different scores in the AQA task arise from expert assessment uncertainty. Therefore, we design a CVAE-based module to encode this uncertainty. Figure~\ref{Fig2} shows the framework of our proposed UD-AQA, which contains two branches: a deterministic prediction branch, and a generative CVAE-based probabilistic latent branch. In the deterministic branch, we follow the existing works to use I3D for clip-level feature extraction and then aggregate several clip-level features into one video-level feature. In the probabilistic branch, we aim to construct a Gaussian distributional latent space to model the uncertainty between the given video and its corresponding judge scores. Features from two branches are connected to produce multiple predictions. We also estimate concrete uncertainty values for further training.

Subsequently, Section~\ref{subsection1} first introduces the extraction of the deterministic video-level features.  Then, the proposed CVAE-based uncertainty modeling process is described in Section~\ref{subsection2}. Section~\ref{subsection3} introduces the estimation of the uncertainty value. Section~\ref{subsection4} further details the objective loss functions. Section~\ref{section5} describes the uncertainty-guided training process.

\subsection{Video-level Feature Extraction} \label{subsection1}
An input video $v_i$ typically  consists of $L$ frames in an AQA dataset $D = \{ v_i,y_i\} _{i = 1}^V$, where $y_i$ is the score label and $V$ denotes the dataset size. We use the I3D~\cite{Carreira2017Quo23} backbone to extract clip-level features following~\cite{Tang2020uncertainty5}, where a sliding window divides one video into $K$ overlapping clips. Each clip is fed to the I3D network to extract clip-level features $\boldsymbol{f}_{c}^k \in {\mathbb{R}^{M}}$, where $k=1,2,..., K$, and $M$ is dimension of the clip features. 

In contrast to the direct average clip features in~\cite{Tang2020uncertainty5}, we devise a weight attention (WA) module to learn the contribution of each clip-level feature to the video-level feature. The WA module comprises four FC layers with Rectified Linear Unit (ReLU)~\cite{agarap2018deep}, except for the last FC layer. We feed each clip-level feature into the WA module to obtain the weights $W \in {\mathbb{R}^{M \times K}}$ and then normalize them ($\bar{W} \in {\mathbb{R}^{M \times K}}$) along the clip dimension by softmax function, which ensures the sum of the clip-level weights to be 1. Subsequently, the video-level feature $\boldsymbol{f}_v \in \mathbb{R}^{M}$ is computed as the weighted sum of clip-level features:

\begin{equation}
  \boldsymbol{f}_v = \sum\limits_{k = 1}^{K} {{{\bar{\boldsymbol{w}}}_k}} \odot \boldsymbol{f}_c^k, 
  \label{VideoFeature}
\end{equation}
where $\odot$ is the element-wise multiplication.

\subsection{CVAE-based Module}\label{subsection2}

CVAE can project high-dimensional data into a low-dimensional latent space. This compact latent space can be characterized by a distribution, such as a Gaussian distribution. By randomly sampling multiple times from this distribution, we can obtain diverse outputs. Therefore, we propose a CVAE-based module in the AQA task, where a random variable sampled from the distribution of low-dimensional latent space can be concatenated with the video representations to predict multiple plausible scores. 

Figure~\ref{Fig2} shows the structure of the CVAE-based module. Specifically, we map the video into a low-dimensional latent space using the designed prior and posterior network. Given the high dimensionality of the video ($L \times 3 \times H \times W$, where $H$ and $W$ denote height and width), we initially extract 3D features from the entire video using four layers of 3D convolutional layers as input data (denoted as $\boldsymbol{X} \in \mathbb{R}^{N}$) for the CVAE-based module. The input channels of the 3D convolutional layers are 3, 16, 32, 64, and the output channels are 16, 32, 64, 128, respectively. Then we use global average pooling to 
ensure that the dimension of $\boldsymbol{X}$ is the same as the final output channel (128). 
It is worth noting that we do not share features between video-level feature extraction and the CVAE-based uncertainty module. This separation is intentional to prevent the training of CVAE from affecting the video feature extraction. Additionally, we design two FC layers to map the low-dimensional latent distribution of the prior and posterior network to a one-dimensional scalar, serving as the estimation of uncertainty.

\textbf{Prior Network.} 
We define ${P_\theta }(\boldsymbol{z}|\boldsymbol{X})$ as the prior network that projects the input variable $\boldsymbol{X} \in {\mathbb{R}^{N}}$ ($N$ is the dimension of the video feature in the CVAE module) to a low-dimensional latent variable $\boldsymbol{z}$. Here,
$\theta$ represents the learnable parameters of the prior network, which includes three FC layers, two ReLU activation function layers, and two dropout layers. We assume the latent space is a $D$-dimensional Gaussian distribution, so the outputs of the prior network contain the $D$-dimensional mean and variance. In our implementation, we output the logarithm of the variance instead of the variance to ensure that the variance is positive. We represent the mean and log variance of the prior network as ${\boldsymbol{\mu}_1},\log ({\boldsymbol{\sigma} _1})$, respectively, where ${\boldsymbol{\mu} _1},\log ({\boldsymbol{\sigma} _1}) \in {\mathbb{R}^D}$.

\textbf{Posterior Network.} 
Another indispensable part of the CVAE model is the posterior network. Formly, we define the posterior network as ${Q_{\boldsymbol{\phi}} }({\boldsymbol{z}}|{\boldsymbol{X}},y)$, which aims to learn the latent variable given the input variable ${\boldsymbol{X}}$ and the condition $y$. Here, ${\boldsymbol{\phi}}$ represents the learnable parameters of the posterior network, ${\boldsymbol{X}}$ is the video feature and $y$ is the corresponding ground-truth score. The posterior network also describes a Gaussian distribution, and we denote the mean and log variance of the Gaussian distribution as ${\boldsymbol{\mu}_2},\log ({\boldsymbol{\sigma}_2})$, respectively, where ${\boldsymbol{\mu}_2},\log ({\boldsymbol{\sigma}_2}) \in {\mathbb{R}^D}$.

During the test process, the condition is not accessible, so we sample from the latent space of the prior network.
To guarantee that the outputs of the prior network in the testing process are reliable enough, we make the output distribution of the prior network and the posterior network as close as possible during the training process. Therefore, Kullback–Leibler (KL) divergence is utilized to measure the distance between these two distributions:
\begin{equation}
    {\mathcal{L}_{\rm latent}} = {D_{\rm KL}}({Q_\phi }({\boldsymbol{z}}|\boldsymbol{X},y)||{P_\theta }({\boldsymbol{z}}|\boldsymbol{X})).
    \label{eq2}
\end{equation}

\subsection{Uncertainty Estimation} \label{subsection3}
After obtaining the $D$-dimensional latent variable $\boldsymbol{z}$, we further map the sampled $\boldsymbol{z}$ to a scalar by two FC layers, representing the logarithm of the uncertainty $u \in \mathbb{R}$. The concrete estimation of uncertainty can indicate the confidence of the predicted results. The training method for uncertainty estimation is similar to that of the latent space. Therefore, we calculate the $L_2$ distance between the uncertainty estimated by the prior network ($u_1$) and the uncertainty estimated by the posterior network ($u_2$) to bring them closer:
\begin{equation}
    {\mathcal{L}_{u}} = 
    \left\| {{u_1}-{u_2}} \right\|_2.
    \label{eq3}
\end{equation}

\subsection{Loss Function}\label{subsection4}

\textbf{Re-weighted AQA loss.} Since uncertainty can serve as an indicator of the importance of each sample, we design a re-weighted AQA loss to ensure that certain samples have high weights, while uncertain ones have low weights during training. Specifically, we first sample the uncertainty-related features $\boldsymbol{f}_u \in {\mathbb{R}^{D}}$ from the distribution of the latent space ($\boldsymbol{z} \sim \mathcal{N}(\boldsymbol{\mu},\boldsymbol{\sigma}^2)$):
\begin{equation}
\boldsymbol{f}_{u}=\boldsymbol{\mu}+\boldsymbol{\epsilon} \cdot \boldsymbol{\sigma}, \boldsymbol{\epsilon} \sim  \mathcal{N}(0,\textbf{I}),
\end{equation}
where $\epsilon$ represents a random noise sampled from a normal distribution. Here we use $\boldsymbol{\mu}_2$, $\boldsymbol{\sigma}_2$ for the training process, and $\boldsymbol{\mu}_1$, $\boldsymbol{\sigma}_1$ for the test process.
Then the uncertainty-related features $\boldsymbol{f}_u$ are concatenated with extracted video-level features $\boldsymbol{f}_v$ (Eq.~\ref{VideoFeature}) as the final feature representations $\boldsymbol{f} \in {\mathbb{R}^{M+D}}$:

\begin{equation}
    \boldsymbol{f} = {\texttt{concat}}(\boldsymbol{f}_{v};\boldsymbol{f}_{u}).
\end{equation}

Subsequently, we feed the feature representations to a Multilayer Perceptron (MLP) regressor, including three FC layers with ReLU activation except for the last layer. This process generates the predictions $\hat y$:
\begin{equation}
    {\hat y}={\varphi_{\rm mlp}(
    \boldsymbol{f}}),
\end{equation}
where $\varphi_{\rm mlp}$ is the parameters of the MLP regressor.

Given that the learned uncertainty serves as a reliable indicator for sample weights, we integrate it into the AQA regression loss to enhance performance. The re-weighted AQA loss is defined as follows:
\begin{equation}
    {\mathcal{L}_{\rm RAQA}} = {e^{ - u}}\left\| {\hat y - y} \right\|_2 + u,
    \label{RAQA}
\end{equation}
where we utilize $e^{-u}$ as the weight for the AQA regression loss. The intuition behind this choice lies in the lower reliability of training samples with higher uncertainty, justifying a reduced weight for the regression loss. Conversely, a larger weight is assigned to samples with lower uncertainty. The second term $u$ acts as a regularization term, mitigating the adverse effects of excessively high uncertainty. During training, we use the output of the posterior network ($u_2$) for uncertainty estimation ($u$), while during testing, the output of the prior network ($u_1$) is utilized for the same purpose.

\textbf{Total loss.} In summary, the total loss function in our proposed UD-AQA is given by
\begin{equation}
    \mathcal{L} = {\mathcal{L}_{\rm RAQA}} + \alpha {\mathcal{L}_{\rm latent}} + \beta {\mathcal{L}_{u}},
\end{equation}
where $\alpha$ and $\beta$ are the weights of the corresponding loss functions.

\subsection{Uncertainty-guided Training Process} \label{section5}
The concrete estimation of uncertainty can play a guiding role in the training process. Curriculum learning suggests that prioritizing easy samples is beneficial for training the model. Thus in this paper, we propose a novel uncertainty-guided training strategy guided by uncertainty to arrange the training order based on uncertainty levels. This approach is adopted because the model exhibits greater confidence in samples with lower uncertainty. Prioritizing lower uncertainty samples facilitates the model in acquiring more accurate information.

Specifically, we begin by recording the predicted uncertainty for each sample after the first epoch and then sort the samples by uncertainty in ascending order. Subsequently, starting from the second epoch, we utilize $p\%$ of the data for training. Once the network converges on the current set of samples, we increment the data utilization by $a\%$ until the entire dataset is incorporated. Throughout the training process, we dynamically update the uncertainty-based sample order obtained from the previous epoch. The convergence criterion is defined as follows: no improvement in performance after three consecutive epochs. Following the approach in~\cite{wang2021survey}, the initial percentage $p$ is set to 40\%, and the incremental step $a$ is set to 20\%.

\section{Experiment}
\subsection{Datasets}
We conduct experiments on three popular AQA datasets, \ie MTL-AQA~\cite{Parmar2019what4}, FineDiving~\cite{xu2022finediving}, and JIGSAWS dataset~\cite{gao2014jhu40}. These datasets contain multiple scores for each action. 

\textbf{MTL-AQA (Multitask AQA) dataset.} The MTL-AQA dataset~\cite{Parmar2019what4} is released for multi-task learning. The motivation is the performance of the AQA task can be improved by learning the descriptions and the details of actions. This dataset only focuses on diving action due to its popularity and promising results. MTL-AQA contains 1412 fine-grained samples. The data is collected from 16 events, including individual or synchronous, 3m Springboard or 10m Platform, and male or female diving. The annotated Labels include AQA scores, action classes, difficulty, and details of actions such as the number of twists and caption. Seven judge scores, along with final scores, are recorded as AQA scores. The final scores are calculated by multiplying the sum of the third to the fifth sorted judge scores with the difficulty level. For our experiments, we split the dataset into 1059 training samples and 353 test samples, following the previous approaches in~\cite{Parmar2019what4,Tang2020uncertainty5,yu2021group42}.

\textbf{FineDiving dataset.} The FineDiving dataset~\cite{xu2022finediving} is collected for analyzing fine-grained sports action to improve the athlete's competitive skills. Each video contains two-level semantic structures, namely action types and sub-action types. Sub-action types represent specific performed sub-actions, such as forward and reverse and an action type is constituted by a combination of these sub-action types. FineDiving dataset includes a total of 3000 video samples, covering 52 action types, 29 sub-action types, and 23 difficulty degree types. The videos depict diving events sourced from notable competitions, including the Olympics, World Cup, World Championships, and European Aquatics Championships on YouTube. The video diving scores are labeled as the product of judge scores and difficulty. The number of judge scores varies, and for consistency, only the three middle valid judge scores are retained. In the experiments, 2251 video samples are utilized for training, while 749 samples are reserved for testing.

\textbf{JIGSAWS dataset.} The JIGSAWS dataset~\cite{gao2014jhu40} involves typical surgical skills curricula videos, including 28 sequences of Suturing (SU), 39 sequences of Knot-Tying (KT), and 35 sequences of Needle-Passing (NP). Each technical skill is annotated based on six elements: respect for tissue, suture, time and motion, flow of operation, overall performance, and quality of final product. Ratings are assigned on a five-point scale for each element, with higher ratings indicating better performance. The final score is derived as the sum of these six sub-scores. We conduct a four-fold cross-validation experiment, with splits consistent with those in~\cite{Tang2020uncertainty5,yu2021group42}.

\subsection{Metrics}
Consistent with prior research~\cite{Tang2020uncertainty5,yu2021group42}, we evaluate the performance using Spearman’s rank correlation ($\rho$) and relative L2-distance. Spearman’s rank correlation is chosen because AQA focuses on the ranking within the competition. The $\rho$ ranges from -1 to +1, indicating the presence and strength of a linear relationship. The higher $\rho$ signifies a more similar rank between the two variables.  The definition of $\rho$ is provided below:
\begin{equation}
   \rho  = \frac{{\sum\nolimits_i {({p_i} - \bar p)} ({q_i} - \bar q)}}{{\sqrt {\sum\nolimits_i {{{({p_i} - \bar p)}^2}\sum\nolimits_i {{{({q_i} - \bar q)}^2}} } } }}, 
   \label{Eq9}
\end{equation}
where $q$ and $\bar q$ denote the ranking of labels and the average ranking of the label set, respectively. Similarly, $p$ and $\bar p $ denote the ranking of predicted results and the average ranking of the prediction sets. 

Given the presence of multiple actions in the JIGSAWS dataset, we calculate the average performance ($\bar \rho$) using the average Fisher’s $z$ score ($\bar z$)~\cite{Parmar2019action7}. The $\bar z$ score is computed from Spearman's rank correlation for each action ($\rho$).

Relative L2-distance (R-$\ell_2$) is introduced by Yu \textit{et al.}~\cite{yu2021group42} as a metric to quantify numerical differences. It is a variant of the L2 distance that takes into account the kinds of actions, acknowledging that different actions may have distinct score ranges. Given the highest and lowest scores $y_{max}$ and $y_{min}$ of the $v$-th action, R-$\ell_2$ is defined as:
\begin{equation}
\mathrm{R}\textrm{-}\ell_{2}=\frac{1}{V} \sum_{v=1}^{V}\left(\frac{\left|y_{v}-\hat{y}_{v}\right|}{y_{\max }-y_{\min }}\right)^{2},
\end{equation}
where $y_v$ and $\hat{y}_v$ represent the ground-truth score and the corresponding prediction, respectively.

\subsection{Implementation Details}
We employ PyTorch framework~\cite{Paszke2017automatic27} as the deep learning framework to implement UD-AQA. For comparison, we follow~\cite{Tang2020uncertainty5} to keep the structure of both the I3D backbone and the regressor the same to explore the effect of the uncertainty module. The I3D model used in clip-level feature extraction is pre-trained on the Kinetics dataset~\cite{Kay2017the25}. The frame number $L$ is 103 for the MTL-AQA and FineDiving datasets, and 160 for the JIGSAWS dataset. The video feature dimension is $M=1024$, and the number of clips is $K=10$. All parameters are updated using the Adam optimizer~\cite{Kingma2014Adam28}, and the weight decay is set to 1e-5. 
The learning rate is 1e-4 for the MTL-AQA and FineDiving datasets, and 1e-3 for the JIGSAWS dataset, respectively.

For training the CVAE-based module, we set the dimension of the input video feature as $N=128$, and the dimension of latent space as $D=6$. 
We set the weight $\alpha$ to 1 and the weight $\beta$ to 10. The number of training epochs is set to 200 for the MTL-AQA and FineDiving datasets, and 80 for the JIGSAWS dataset.

\begin{table}[ht]
  \centering
\begin{threeparttable}
  \caption{The results on the MTL-AQA dataset.}
  \setlength{\tabcolsep}{4mm}{
    \begin{tabular}{ccc}
    \toprule
    Method & Sp. Corr $\uparrow$& R-$\ell_2$($\times 100$) $\downarrow$ \\
    \midrule
    Pose+DCT~\cite{Pirsiavash2014assessing1} & \multicolumn{1}{c}{0.2682}&\multicolumn{1}{c}{-}  \\
    C3D-SVR~\cite{parmar2017learning2} & \multicolumn{1}{c}{0.7716}&\multicolumn{1}{c}{-} \\
    C3D-LSTM~\cite{parmar2017learning2} & \multicolumn{1}{c}{0.8489}  &\multicolumn{1}{c}{-}\\
    C3D-AVG-MTL~\cite{Parmar2019what4} & \multicolumn{1}{c}{0.9044}&\multicolumn{1}{c}{-} \\
    Regression~\cite{Tang2020uncertainty5}&{0.8905}&\multicolumn{1}{c}{-}
    \\
    USDL~\cite{Tang2020uncertainty5}&{0.9066}&\multicolumn{1}{c}{-}  \\
    MUSDL~\cite{Tang2020uncertainty5} & \multicolumn{1}{c}{0.9273}&\multicolumn{1}{c}{-}\\
    PLCN~\cite{bai2022action} & \multicolumn{1}{c}{0.9230}&\multicolumn{1}{c}{-}
    \\
    CoRe~\cite{yu2021group42}&{0.9512} &\multicolumn{1}{c}{0.2600}
    \\
    \hline
    Ours  & \multicolumn{1}{c}{\textbf{0.9545}}&\multicolumn{1}{c}{\textbf{0.2590}}  \\
    \bottomrule
    \end{tabular}}
  \label{table1}
  \end{threeparttable}
\end{table}

\subsection{Results on MTL-AQA Dataset}
In Table~\ref{table1}, we present the experimental results of our proposed UD-AQA and comparisons with existing state-of-the-art approaches built upon the similar CNN-based backbones on the MTL-AQA dataset. Our approach demonstrates new state-of-the-art performance, particularly excelling in terms of Spearman's rank correlation. Notably, our approach surpasses MUSDL~\cite{Tang2020uncertainty5} with a substantial performance gain of $2.72\%$. Additionally, when compared to the GART-based CoRe~\cite{yu2021group42}, we outperform it by $0.33\%$, utilizing only a common regression backbone. Furthermore, in measuring R-$\ell_2$ distance, we achieve slightly better performance than CoRe (0.259 vs 0.260 on MTL-AQA). This improvement is noteworthy as UD-AQA employs a simple regression head, while CoRe incorporates a more complex coarse-to-fine multi-layer regression tree.

More importantly, the previous methods~\cite{Tang2020uncertainty5} simply convert a single decision head to multiple heads to generate different outputs, which is not scalable when there is a varying number of judge scores available. CoRe~\cite{yu2021group42} needs to choose 10 exemplars for voting one predicted score, so it only provides one final score rather than each judge score due to slow inference time. Instead, our work focuses on generating multiple and diverse outputs, which can be easily obtained by sampling $t$ times during the test process. 

\subsection{Results on FineDiving Dataset}
We also conduct experiments on the new public FineDiving dataset~\cite{xu2022finediving}. The FineDiving dataset also contains multiple judge scores, making it suitable for our design. As shown in Table~\ref{table2}, the performance gain of our proposed UD-AQA becomes more apparent due to the FineDiving dataset having a larger number of training samples. Guided by the estimated uncertainty, our proposed UD-AQA even surpasses TSA~\cite{xu2022finediving} with 15 exemplars 1.37\% in Spearman's rank correlation and 0.0434 in terms of R-$\ell_2$.  UD-AQA also outperforms the second best method~\cite{lian2023improving} by 1.19\% (Spearman's rank correlation) and 0.0319 (R-$\ell_2$). Considering that ~\cite{lian2023improving}
divides the action into three sub-stages and designs an across-staged temporal reasoning module, our UD-AQA, on the contrary, does not involve complicated architectural design, demonstrating the effectiveness of the CVAE-based uncertainty estimation.

\begin{table}[ht]
  \centering
\begin{threeparttable}
  \caption{The results on the FineDiving dataset.}
  \setlength{\tabcolsep}{4mm}{
    \begin{tabular}{ccc}
    \toprule
    Method & Sp. Corr $\uparrow$& R-$\ell_2$($\times 100$) $\downarrow$ \\
    \midrule
    TSA-1~\cite{xu2022finediving} & 0.9085&0.4020\\
    TSA-5~\cite{xu2022finediving} & 0.9154&0.3658\\
    TSA-10~\cite{xu2022finediving} & 0.9203 &0.3420\\
    TSA-15~\cite{xu2022finediving} & 0.9204 & 0.3419\\
    Lian \textit{et al.}~\cite{lian2023improving} &0.9222&0.3304
    \\
    \hline
    Ours  & \multicolumn{1}{c}{\textbf{0.9341}}&\multicolumn{1}{c}{\textbf{0.2985}}  \\
    \bottomrule
    \end{tabular}}
  \label{table2}%
  \begin{tablenotes}
		\item $N$ in TSA-$N$ denotes the number of exemplars for each test sample.
	\end{tablenotes}
  \end{threeparttable}
\end{table}

\subsection{Results on the JIGSAWS Dataset}
\begin{table}[ht]
  \centering
  \caption{The results on the JIGSAWS dataset.}
  \setlength{\tabcolsep}{3mm}{
    \begin{tabular}{ccccc}
    \toprule
    Method & \multicolumn{1}{c}{S} & \multicolumn{1}{c}{NP} & \multicolumn{1}{c}{KT} & \multicolumn{1}{c}{Avg.Corr. $\uparrow$} \\
    \midrule
    ST-GCN~\cite{Yan2018Spatial30}& 0.31  & 0.39  & 0.58  & 0.43 \\
    TSN~\cite{Pan2019Action8,wang2016temporal36}   & 0.34  & 0.23  & 0.72  & 0.46 \\
    JRG~\cite{Pan2019Action8}  & 0.36  & 0.54  & 0.75  & 0.57
    \\
    USDL~\cite{Tang2020uncertainty5} & 0.64  & 0.63  & 0.61  & 0.63\\
    MUSDL~\cite{Tang2020uncertainty5} & 0.71  & 0.69  & 0.71  & 0.70\\
    CoRe~\cite{yu2021group42} & 0.84  & 0.86  & 0.86  & 0.85 \\
    Ours  & \textbf{0.87}  & \textbf{0.93}  & \textbf{0.86}  & \textbf{0.89} \\
    \midrule\midrule
     R-$\ell_2$($\times 100$)& S & NP & KT & Avg. $\downarrow$\\
    \midrule
    CoRe&5.055&5.688&\textbf{2.927}&4.556\\
    Ours&\textbf{3.444}&\textbf{4.076}&5.469&\textbf{4.330}\\
    \bottomrule
    \end{tabular}}
  \label{table3}%
  \vspace{-0.1cm}
\end{table}

\begin{table}[ht]
 \centering
 \caption{The ablation study on the MTL-AQA dataset.}
\begin{tabular}{c|cccc|c}
    \toprule
    $\#$ & WA  & \makecell[c]{CVAE} &\makecell[c]{UE} &\makecell[c]{Order}& \multicolumn{1}{c}{Sp. Corr} \\
    \midrule
    1& \multicolumn{1}{c}{} & \multicolumn{1}{c}{} & & \multicolumn{1}{c|}{} & 0.9364  \\
    2& \multicolumn{1}{c}{} & \checkmark   & &\multicolumn{1}{c|}{} & 0.9465 \\
    3& \checkmark & \checkmark &     & & 0.9478    \\
    4& \checkmark & \checkmark    &\checkmark& &     0.9535 
    \\
    5& \checkmark     & \checkmark & \checkmark &   \checkmark  &   0.9545    \\
    \bottomrule
    \end{tabular}
  \label{table4}%
\end{table}

\begin{table*}[t]
  \caption{{The results with different network designs on the MTL-AQA dataset.}}
  \centering
  \setlength{\tabcolsep}{3mm}{
    \begin{tabular}{c|cc|cc|cc}
    \hline
    \multirow{2}*{Model} & \multicolumn{2}{c}{Features}\vline&\multicolumn{2}{c}{Regressor}\vline&\multirow{2}*{Sp. Corr $\uparrow$} & \multirow{2}* {R-$\ell_2$($\times 100$) $\downarrow$ }\\
    \cline{2-5}
    ~&Separate Features & Shared Features&One Regressor& Two Regressors&~&~ \\
    \hline
    1&\checkmark&&\checkmark& &{{0.9545}}&\multicolumn{1}{c}{{0.2590}}  \\
    2& & \checkmark&\checkmark&  & \multicolumn{1}{c}{{0.9539}}&\multicolumn{1}{c}{{0.3329}}  \\
    \hline
    3 & \checkmark& & & \checkmark& \multicolumn{1}{c}{{0.9514}}&\multicolumn{1}{c}{{0.3222}}  \\
    4 & &\checkmark & & \checkmark& \multicolumn{1}{c}{{0.9530}}&\multicolumn{1}{c}{{0.3976}}  \\
    \hline
    \end{tabular}}
 \label{table5}
\end{table*}

In addition to the Olympic Events, we also conduct experiments on the surgical skills assessment dataset JIGSAWS. We first select the best model for each fold like CoRe~\cite{yu2021group42}, then we average the results of four folds. The results of our proposed UD-AQA and other classic methods are presented in Table~\ref{table3}.

From the results, we can observe that our UD-AQA achieves the best result in all activities (0.87 for S, 0.93 for NP, and 0.86 for KT), and establishes a new state-of-the-art average correlation (0.89).  The performance gain is significant, surpassing the second-best CoRe~\cite{yu2021group42} by a large margin of 4\%. In terms of R-$\ell\_2$, our UD-AQA also outperforms CoRe by 0.226. These results suggest that the introduction of uncertainty can significantly benefit the AQA task.

\subsection{Ablation Study}
UD-AQA contains the following effective strategies: the weight attention (WA) module for video-level features, the CVAE-based module (CVAE for short) for sampling uncertainty-related features from the distribution of latent space, the uncertainty estimation (UE for short) for re-weighting AQA loss, and the strategy of uncertainty-guided training order (Order for short). We conduct ablation studies on the MTL-AQA dataset to explore the role of each component. Note that model \#1 serves as the baseline, which only contains the I3D backbone and MLP layers. Table~\ref{table4} displays the ablation results. We also conduct ablation studies on the selection of the latent space dimension, and the results are shown in Figure~\ref{Fig3}. Furthermore, we test different design variants to evaluate the effectiveness of our network design shown in Table~\ref{table5}.

\subsubsection{Effects of weight attention module} The WA module alone cannot produce diverse results, so we integrate it with the CVAE-based module to showcase its effectiveness. A comparison between models \#2 and \#3 comprehensively illustrates the advantage of the weight attention strategy. Directly averaging the clip-level features is too simple. In contrast, we capture more rich video-level features by learning the contribution of each clip-level feature.

\subsubsection{Effects of the CVAE-based module} We use the CVAE-based module to sample from the distribution of latent space for generating multiple scores, and design a latent loss $\mathcal{L}_{\rm latent}$ (Eq.~\ref{eq2}) to constrain the distribution estimation of the latent space. For verification, we add a CVAE-based module and a latent loss (\#2) based on the baseline (\#1). Compared with the baseline, Spearman's rank coefficient increases by 1.01\%, demonstrating the effectiveness of the CVAE-based module. Importantly, the CVAE-based module allows us to generate multiple possible outputs by sampling from the latent space. Given its crucial role in achieving diverse outputs, we retain the CVAE-based module in all subsequent models \#3-\#5.

\subsubsection{Effects of uncertainty estimation} 
We output a scalar value for uncertainty estimation by constraining $\mathcal{L}_{u}$ (Eq.~\ref{eq3}) and further use this scalar to re-weight the AQA loss (Eq.~\ref{RAQA}). 
To validate the effect of uncertainty estimation, model \#4 adds $\mathcal{L}_{u}$ (Eq.~\ref{eq3}) and replaces the regular AQA loss with the re-weighted AQA loss (Eq.~\ref{RAQA}) compared with model \#3. The results show that the uncertainty loss and the re-weighted AQA loss play a positive role. 

\subsubsection{Effects of the strategy of uncertainty-guided training order} Based on the predicted uncertainty, we order the training samples from low uncertainty to high uncertainty. This scheme enforces the model to learn the information from confident to uncertain, making the training process more effective compared to a random order. A comparison between models \#4 and \#5 serves as a comprehensive validation of this training strategy.

\begin{figure}[ht]
\centering
 \includegraphics[scale=0.5]{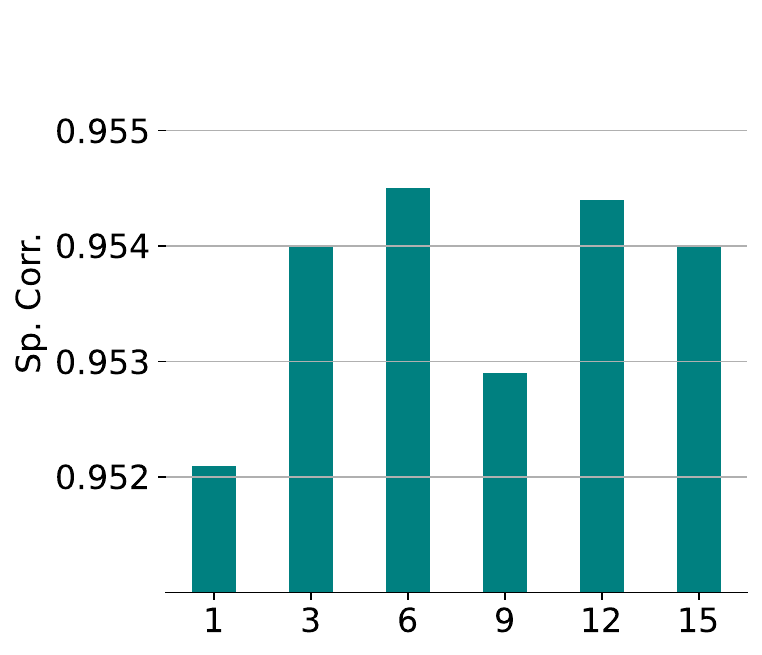}
 \caption{Effects of the dimension of latent space.}
 \label{Fig3}
\end{figure}

\subsubsection{Effects of the latent space dimension in the CVAE-based module} The dimension of latent space is a vital hyperparameter determining how much the latent space affects the fused feature representations. Following the previous work~\cite{Kohl2018probabilistic20}, we range the latent space dimension $D$ in $[1,3,6,9,12,15]$ and keep other variables optimal. We can find that the suitable dimension for latent space is 6 as shown in Figure~\ref{Fig3}. 
Continuously increasing the dimension will not yield additional benefits.

\begin{table*}[ht]
  \centering
  \caption{Comparisons of performance with existing methods
  on the AQA-7 dataset.}
    \begin{tabular}{ccccccccc}
    \toprule
    \multicolumn{1}{c}{} & \multicolumn{1}{c}{Diving} & \multicolumn{1}{c}{Gym Vault} & \multicolumn{1}{c}{Skiing} & \multicolumn{1}{c}{Snowboard} & \multicolumn{1}{c}{Sync. 3m} & \multicolumn{1}{c}{Sync. 10m} & \multicolumn{1}{c}{Avg. Corr} \\
    \midrule
    Pose+DCT~\cite{Pirsiavash2014assessing1} & 0.5300  & \multicolumn{1}{c}{-} & \multicolumn{1}{c}{-} &       & \multicolumn{1}{c}{-} & \multicolumn{1}{c}{-} & \multicolumn{1}{c}{-} \\
    ST-GCN~\cite{Yan2018Spatial30} & 0.3286 & 0.5770 & 0.1681 & 0.1234 & 0.6600  & 0.6483 & 0.4433 \\
    C3D-LSTM~\cite{parmar2017learning2} & 0.6047 & 0.5636 & 0.4593 & 0.5029 & 0.7912 & 0.6927 & 0.6165 \\
    C3D-SVR~\cite{parmar2017learning2} & 0.7902 & 0.6824 & 0.5209 & 0.4006 & 0.5937 & 0.9120 & 0.6937 \\
    All-action C3D-LSTM~\cite{Parmar2019action7} & 0.6177 & 0.6746 & 0.4955 & 0.3648 & 0.8410 & 0.7343 & 0.6478 \\
    JRG~\cite{Pan2019Action8} & 0.7630 & 0.7358 & 0.6006 & 0.5405 & 0.9013 & 0.9254 & 0.7849 \\
    AIM~\cite{Gao2020An29} & 0.7419 & 0.7296 & 0.5890 & 0.4960 & 0.9298 & 0.9043 & 0.7789 \\
    Regression~\cite{Tang2020uncertainty5} & 0.7438 & 0.7342 & 0.5190 & 0.5103 & 0.8915 & 0.8703 & 0.7472 \\
    USDL~\cite{Tang2020uncertainty5} & 0.8099 & 0.7570 & 0.6538 & 0.7109 & 0.9166 & 0.8878 & 0.8102 \\
    EAGLE-Eye~\cite{Nekoui2021EAGLE6} & 0.8331 & 0.7411 & 0.6635 & 0.6447 & 0.9143 & 0.9158 & 0.8140 \\
    CoRe~\cite{yu2021group42}&0.8824&0.7746&0.7115&0.6624&0.9442&0.9078&0.8401\\
    \midrule
    Ours  & 0.8532 & 0.7663 & 0.6836 & 0.5596 & 0.9281 & 0.9438 & 0.8318 \\
    \bottomrule
    \end{tabular}%
  \label{tab:table6}%
\end{table*}%

\subsubsection{Different design variants}
In this section, we summarize and evaluate the experimental designs we used. First, for designing the feature extraction of two branches, we have two choices: separate or shared feature extraction. ``Separate Features'' scheme means the feature extraction of the deterministic branch and the latent branch is independent to generate diverse scores. For the deterministic branch, I3D with clip and WA module in the video-level feature extraction module extracts better video-level features for fair comparisons with other methods. For the latent branch, I3D with the entire video in the CVAE-based module directly extracts the whole video-level features for less computational cost. ``Shared features'' means that we utilize the extracted video-level feature $\boldsymbol{f}_v$ in the deterministic branch as the input $\boldsymbol{X}$ of the latent branch. From Table~\ref{table5}, we can find separate feature strategy outperforms the shared feature strategy, with particularly significant gains, especially in the R-$\ell_2$ metric.

After getting the feature representations, we need to feed them into a regressor for generating scores. In our UD-ADA, we connect the video-level features extracted from the deterministic branch with the features sampled from the latent branch and then send them into a single regressor to obtain judge scores (denoted as ``One Regressor''). We can also use another optional scheme, denoted as ``Two Regressors'', where the video-level features extracted from the deterministic branch and the features sampled from the latent branch are respectively fed into two separate regressors. The supervised target of the deterministic features is the mean of multiple judge scores, and the supervised target of the latent features is the offset between the sampled judge score and the averaged score. This way can help the latent space learn the ambiguity of the multiple plausible judge scores. Table~\ref{table5} shows the results with different network designs. We can see that the combination of separate feature extraction and a single regressor for two branches achieves the best performance, demonstrating the effectiveness of our network design.

\subsection{Further Analysis} \label{further}
Although our proposed UD-AQA is motivated by generating multiple predicted scores, it should be emphasized that the requirement for multiple ground-truth labels is not necessary. UD-AQA can also work for dealing with single ground-truth labels, but it is better with multiple labels, which can improve the performance from the diversity of ground-truth labels. To verify this, we conduct an experiment on the AQA-7 dataset with one single ground-truth label. We further visualize the estimated uncertainty and discuss the computational cost. 

\subsubsection{The results on the AQA-7 dataset~\cite{Parmar2019action7}} UD-AQA can also deal with the dataset with only one ground-truth label, but multiple labels are more appreciated due to the introduction of more diverse scores. To evaluate the performance on the dataset with a single ground-truth label, we report the performance on the AQA-7 dataset~\cite{Parmar2019action7}, where only final scores are provided. 

Specifically, the AQA-7 dataset captures seven actions from Summer and Winter Olympics. The dataset comprises 370 single diving 10m platform sequences, 176 gymnastic vault sequences, 175 big air skiing sequences, 206 big air snowboarding sequences, 88 synchronous diving-3m Springboard sequences, 91 synchronous diving-10m platform sequences, and 83 trampoline sequences. The average frames of the trampoline sequences are 634, which is much larger than the average frames of about 100 frames for diving and other actions, so we follow previous works~\cite{Parmar2019action7} to discard trampoline sequences. Only the final scores labels are available while the multiple judge scores are unknown in the AQA-7 dataset. The score range is varied for different actions so normalization is needed. 803 sequences belong to the training set and 303 sequences are for the testing.

Table~\ref{tab:table6} summarizes the performance comparison on the AQA-7 dataset. We can see that the average result on the AQA-7 dataset is 0.8318, which is 2.16\% higher than multi-head USDL~\cite{Tang2020uncertainty5}. Compared with other methods, our UD-AQA achieves the second-best result, being only lower than the best CoRe. The potential reason is that our focus is to generate multiple scores, so UD-AQA only uses a simple regression head, while CoRe designs a coarse-to-fine multi-layer regression tree.  If multiple labels are available, the performance can be improved further by involving more diversity. 

\subsubsection{Computational cost} Compared to the baseline model which occupies 19.21GB GPU with a batch size of 4 during training, our UD-AQA requires 21.15GB GPU under the same settings. This increase is attributed to the introduction of the uncertainty module. However, it is important to note that our UD-AQA can generate multiple scores with only a marginal increase in computational cost. This increase remains affordable for real-world applications.

\section{Conclusion}
In this paper, we propose a novel probabilistic model called the Uncertainty-Driven AQA model (UD-AQA), which can encode the uncertainty to generate multiple scores in the action quality assessment task. The introduction of uncertainty enables diverse outputs possible, which is vital for real application scenarios. Moreover, the estimation of uncertainty indicates the confidence of the predicted results, which can re-weight the importance of each sample for training and further guide the training order of samples. Our method enhances scalability and efficiency for multi-judge datasets by considering uncertainty. This allows for the easy generation of more judge scores by increasing the number of sampling times. The results obtained on MTL-AQA, JIGSAWS, and FineDiving datasets highlight the effectiveness of the proposed method.

\bibliographystyle{IEEEtran}
\bibliography{egbib}

\end{document}